%% file: main.tex
\documentclass[10pt,twocolumn,letterpaper]{article}
\usepackage[applications]{wacv}
\usepackage{epsfig,graphicx,amsmath,amssymb,booktabs,xcolor,multirow}
\usepackage{newtxtext}
\definecolor{wacvblue}{rgb}{0.21,0.49,0.74}
\usepackage[breaklinks,colorlinks,allcolors=wacvblue]{hyperref}  %
\hypersetup{pdftitle={How reliable is an embedding of the seabed? Repeat-look measurement of frozen encoders on synthetic aperture sonar},pdfauthor={Harold Horsley, Oscar Pizarro, Hayat Rajani, Morten Omholt Alver}}
\def\wacvPaperID{*****}
\def\confName{WACV}
\def\confYear{2027}
\title{How reliable is an embedding of the seabed? Repeat-look measurement of frozen encoders on synthetic aperture sonar}
\author{Harold Horsley$^{1}$ \qquad Oscar Pizarro$^{1}$ \qquad Hayat Rajani$^{2}$ \qquad Morten Omholt Alver$^{1}$\\
$^{1}$Norwegian University of Science and Technology \qquad $^{2}$University of Girona}

\begin{document}
\maketitle
\input{sec_abstract}
\input{sec_intro}
\input{sec_related}

\input{sec_setup}

\input{sec_results}

\input{sec_discussion}
{\small\bibliographystyle{ieeenat_fullname}\bibliography{refs_zotero}}
\clearpage
\setcounter{section}{0}\setcounter{figure}{0}\setcounter{table}{0}\setcounter{equation}{0}
\renewcommand{\thesection}{S\arabic{section}}
\renewcommand{\thefigure}{S\arabic{figure}}
\renewcommand{\thetable}{S\arabic{table}}
\renewcommand{\theequation}{S\arabic{equation}}
\twocolumn[{\centering\Large\bfseries Supplementary material\par\vspace{4pt}
\large How reliable is an embedding of the seabed? Repeat-look measurement of frozen encoders on synthetic aperture sonar\par\vspace{14pt}}]
\input{sec_supp}
\end{document}

%% file: sec_abstract.tex
\begin{abstract}
Pretrained encoders from self-supervised representation learning are increasingly applied to sonar
imagery of the seafloor, where they are evaluated with protocols that take one embedding per target. A
target patch of seabed, by contrast, is re-imaged from an alternate look on overlapping survey passes, so each
embedding is one measurement of that target. Standard protocols cannot say how
consistent those measurements are, or how much of their spread belongs to the patch and how much to the
acquisition geometry. We treat this as a measurement-reliability problem. Using a synthetic-aperture sonar (SAS) survey, we measure the repeat-look reliability of embeddings from a frozen natural-image
encoder (DINOv3 ViT-S/16). Standard single-look metrics, effective rank, clustering and a linear probe
for coral, are all stable to the look, and the probe performs well (AUC 0.93). Yet the repeat-look
reliability of the same embeddings is poor, with a distance-based intraclass correlation (dbICC) of
0.20 and nearest neighbours and clusters that barely agree across looks. This contradiction resolves as we
find the probe learns a stable direction, whose ICC of 0.60 is 3.0$\times$ that of the
full embedding.
Variance components, calculated from generalizability theory, attribute under a tenth of the
within-cell spread to look acquisition geometry. The rest is cell-specific and cannot be separated without
repeat passes at matched geometry. A sonar-pretrained encoder (BenthicDINO) and a classical SAS texture
descriptor show the same pattern. For this dataset and the tested representations, uses that depend on
patch identity fail, such as cross-pass comparison and retrieval, while uses at the level of habitat type
(e.g.\ coral classification) do not. Because the standard metrics did not detect this, we argue that
repeat-look reliability should accompany any evaluation of representations of the seafloor.
\end{abstract}

%% file: sec_intro.tex
\section{Introduction}\label{sec:intro}

\begin{figure}[t]
\centering
\includegraphics[width=0.8\linewidth]{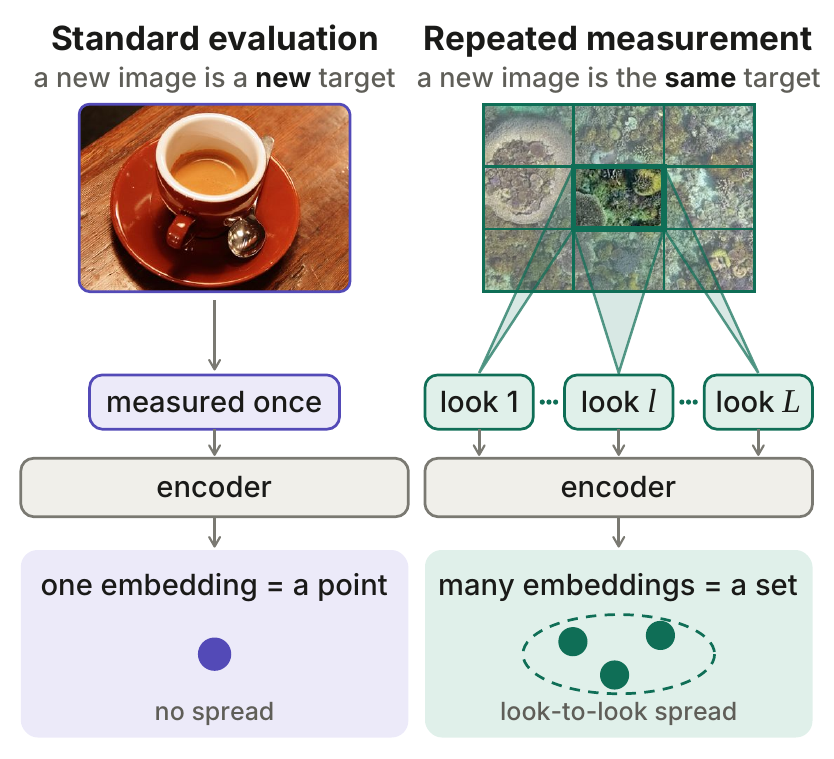}
\caption{Schematic of the measurement process that yields an embedding, in standard evaluation (left)
and repeated measurement of one target (right).}
\label{fig:identity}
\end{figure}

Standard evaluation of pretrained representations (e.g.\ linear probes on frozen features \cite{kolesnikov2019revisiting},
$k$-nearest-neighbour classification \cite{wu2018unsupervised}, geometric stability \cite{geomstability2026}) typically takes one embedding per target, a single point in embedding space.
When the target is a physical area imaged many times, each embedding is
one measurement of that target (Figure~\ref{fig:identity}). Unless the representation has collapsed,
these embeddings differ, and the target is represented by the set of them. Standard evaluation does not
characterise this set. While retrieval and place recognition hold repeat images of a scene, they assess
matching and not the reliability of the embeddings \cite{radenovic2018revisiting,berton2022benchmark} (Section~\ref{sec:related}). We therefore draw metrics
and methods from other fields to characterise it, and ask what the sources of variation are, how large
each is, what they do to downstream tasks, and how the answers can guide further development.

Whilst self-supervised encoders are typically trained to be invariant to image augmentations, the
\emph{views} of contrastive learning \cite{chen_simple_2020,tian2020makes}, an augmentation cannot easily replicate an alternate \emph{look} \cite{synnes2021speckle}.
A look is a new capture of the scene from a different geometry. Some methods do train with natural positives in the loss,
nearby tiles \cite{jean2019tile2vec,yamada2022geoclr} or repeat looks \cite{manas2021seasonal}, or through an explicit invariance penalty
\cite{hamoda2026benthicdino}, but their embeddings will still vary between looks, and so a way to
evaluate that variation is required.

Here we use repeat looks from the overlapping passes of a synthetic aperture sonar (SAS) survey of a
cold-water coral reef. We first apply a standard suite of evaluation metrics for embedding quality
(i.e.\ effective rank, intrinsic dimension, clustering) and a downstream prediction task (coral
classification with a linear probe). We then apply methods drawn from test--retest reliability, representational similarity
analysis and generalizability theory (Section~\ref{sec:related}) to measure \emph{repeat-look
reliability}, defined here as the share of a score's variance that belongs to the cell, a quantity that depends
on which look-to-look variation is counted as error. Furthermore, we decompose the look-to-look variation into named variation effects and a
cell-specific remainder, and examine what the standard metrics fail to see. Three representations are
compared: a state-of-the-art frozen natural-image encoder (DINOv3), a side-scan sonar (SSS) pretrained
encoder (BenthicDINO) and a classical SAS texture descriptor.

We find that the standard metrics are stable to the choice of look, while the repeat-look reliability
coefficients are low for every representation and lowest on the reef. A linear probe on coral labels nevertheless performs
well, and the single direction it uses is several times more reliable across looks than the embedding
as a whole. The named variation effects explain a small part of the spread, and the rest is
cell-specific.

%% file: sec_related.tex
\section{Related Work}\label{sec:related}

\noindent\textbf{Learned representations on sonar.} Deep learning on sonar imagery \cite{steiniger_survey_2022a,aubard_sonarbased_2025} relies mostly on
supervised pipelines (e.g.\ \cite{sture2018recognition}), often with transfer from natural-image pretraining \cite{huo2020underwater}, and on
self-supervised pretraining followed by fine-tuning \cite{kwon2026minejepa}. Where encoder features are kept frozen, they feed
task-trained heads \cite{sheffield2023selfsupervised,kwon2026minejepa}. Self-supervised sonar encoders exist for
forward-looking sonar \cite{preciado2022selfsupervised} and for side-scan \cite{hamoda2026benthicdino},
the latter the only one with public weights that we found, and neither has been evaluated on real repeat looks.

\noindent\textbf{Reliability in sonar imagery.} Reliability is well quantified in coherent imaging (sonar
and radar), where the coherence of repeat acquisitions measures it \cite{bamler1998synthetic,touzi1999coherence}. In sonar, backscatter can depend strongly on
ensonification azimuth \cite{lurton2018azimuth}, backscatter repeatability is monitored over natural
reference areas \cite{roche2018control}, repeat-pass speckle coherence has been characterised for SAS
\cite{synnes2021speckle}, and cross-line appearance differences of the same patch have been
quantified by descriptor matching \cite{xu2023canonical}. Repeat-pass decorrelation has been quantified for radar
\cite{zebker1992decorrelation}. To our knowledge, these characterisations have not been applied to
learned representations.

\noindent\textbf{Reliability of learned representations.} Representation reliability has mostly been measured
model-side \cite{klabunde2025similarity}, as the similarity of representations learned from different training seeds or
initialisations \cite{li2016convergent,raghu2017svcca,kornblith_similarity_2019,mehrer2020individual} and as robustness to
perturbations of the representation itself \cite{geomstability2026}. Visual place recognition and instance retrieval group
repeat images of a scene but score retrieval success, Recall@$N$, mAP or precision--recall, and not the reliability of
the embeddings \cite{radenovic2018revisiting,berton2022benchmark,zaffar2021vprbench}. Input-side studies, in which the target
itself is re-imaged and reliability measured, are recent and come from medical imaging: test--retest reliability of a CT foundation model's predictions
\cite{pai2024foundation}, and per-dimension variance components and intraclass correlations of brain-MRI
foundation-model embeddings across scanners \cite{navarro2026travelling}.
Representational similarity analysis compares the distance structure of two representations of the
same stimuli \cite{kriegeskorte2008rsa,nili2014toolbox} and has its own reliability across repeated
measurements \cite{walther2016reliability}. The distance-based intraclass correlation extends
test--retest reliability \cite{nunnally1978psychometric} to vector-valued measurements
\cite{xu2021dbicc,shou2013i2c2}. The closest precedent to our setting is the scanner shift of pathology foundation models studied by
\citet{thiringer2026scanner}, where the same tissue section imaged on five scanners gives embeddings
whose distance structure is distorted while downstream accuracy is retained. We found no measurement
of embedding reliability under re-measurement for sonar representations or other applications in the
marine realm.

\noindent\textbf{Generalizability theory (G-theory).} Classical test theory treats a measurement as a true
score plus one undifferentiated error \cite{nunnally1978psychometric}. G-theory
\cite{cronbach1972dependability,brennan1992items,brennan2001generalizability,webb2006gtheory} replaces the single error
with named sources of variation, called facets, whose variance components are estimated with
random-effects analysis-of-variance methods \cite{searle1992variance,brennan2001generalizability}. A
facet can be considered as random (error), or fixed, in which case its interaction with the object of
measurement becomes part of the target \cite[\S3.3]{webb2006gtheory}. G-theory defines a reliability
coefficient for each universe of generalization, that is, for each choice of facet as fixed or random. The theory is standard in psychometrics and is also used for test--retest
reliability in neuroimaging \cite{noble2017influences}. For a vector-valued measurement, summing the one-facet variance components
over dimensions before taking the ratio gives the image intraclass correlation of \citet{shou2013i2c2},
which equals the distance-based ICC under Euclidean distance \cite{xu2021dbicc}. We are not
aware of a G-theory analysis of learned representations. In metrology, the precision of measurements repeated under specified
changed conditions is intermediate precision \cite{jcgm2012vim,iso57253}, the metrological counterpart of our repeat-look
conditions. What G-theory allows is a reliability coefficient defined per use, by choosing which
sources of variation count as error, and the projection of the estimated components to survey designs
through a D-study \cite[ch.~4]{brennan2001generalizability}, \cite{shavelson1989generalizability}. One can then not only ask whether
a representation is reliable but also for what task.

%% file: sec_setup.tex
\section{Methods}\label{sec:methods}

\subsection{Data}\label{sec:data}

\begin{figure}[t]
\centering
\IfFileExists{figures/survey/fig_looks_per_cell.png}{\includegraphics[width=0.82\linewidth]{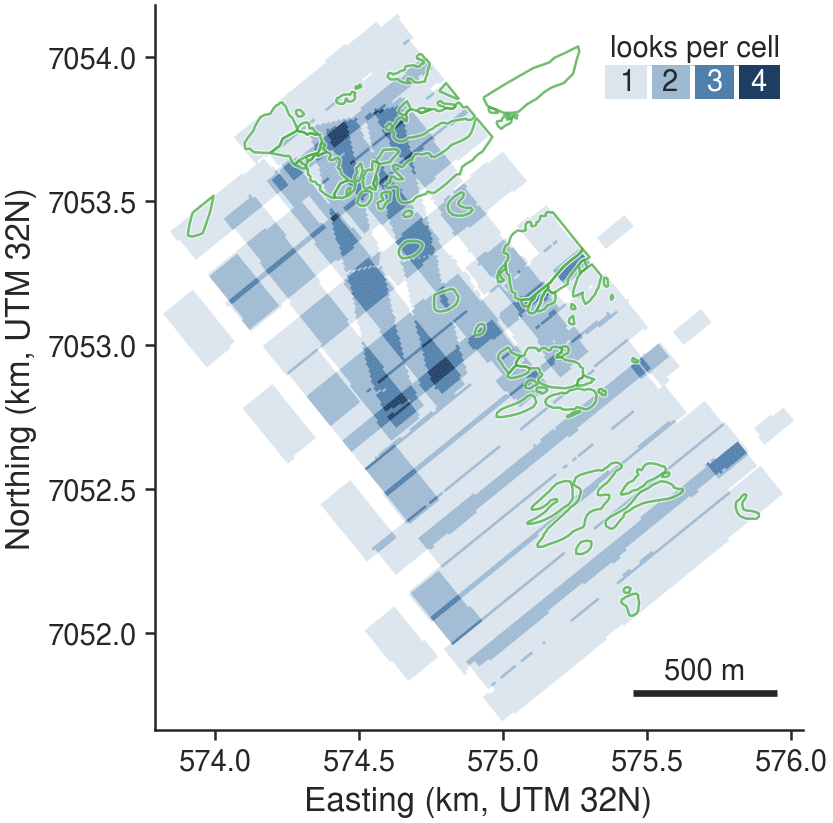}}{\includegraphics[width=\linewidth]{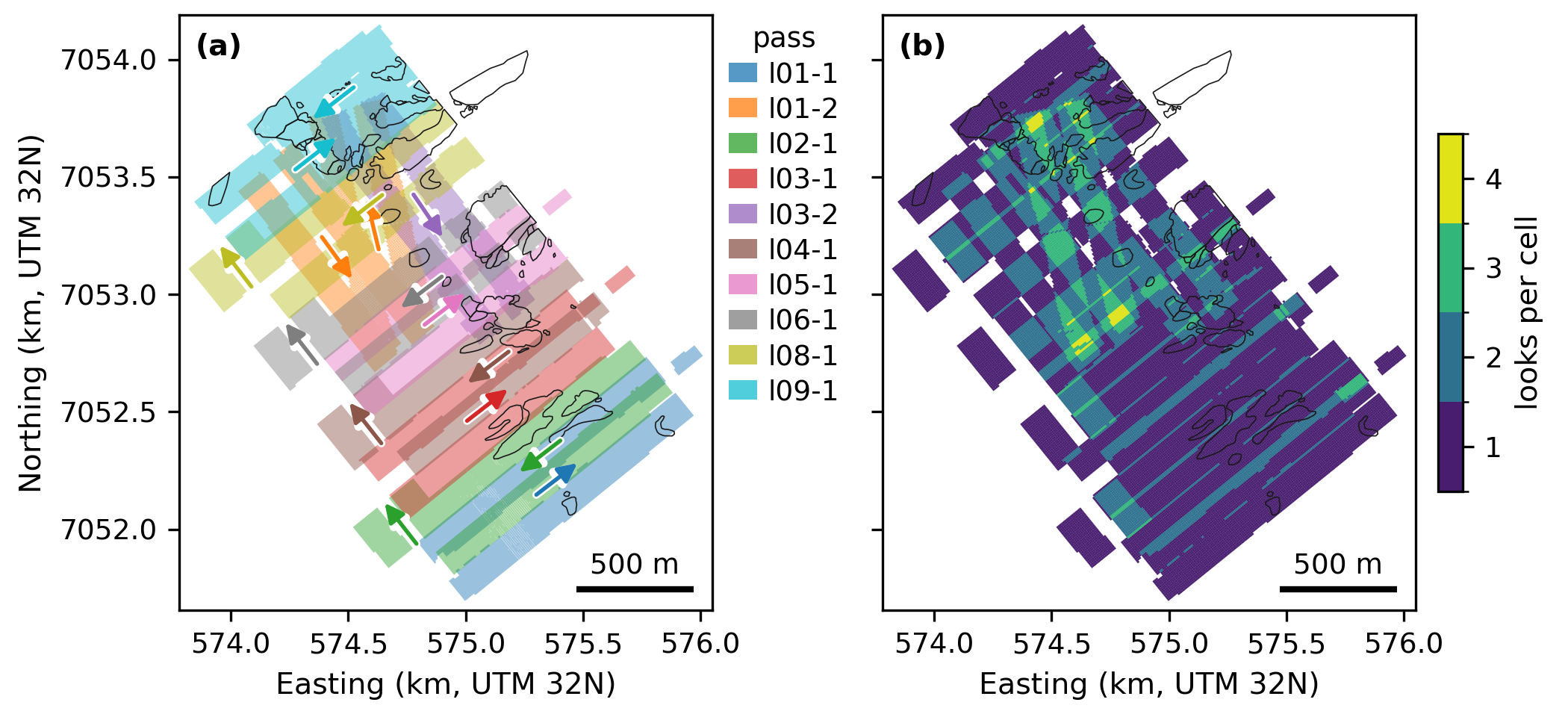}}
\caption{Survey coverage. Number of looks per $8.25$\,m cell over the 2012 Tautra SAS survey, from one
(light) to four (dark). The north-west crossing passes give the cells with three and four looks. Green
outlines mark the 2012 coral-reef polygons.}
\label{fig:survey}
\end{figure}

\noindent\textbf{Survey.}
A synthetic-aperture sonar (SAS) dataset obtained from a Kongsberg HISAS~1030 on the HUGIN-class AUV Hugin HUS over
the Tautra cold-water coral reef (Trondheimsfjord, Norway) in December 2012, at a mean altitude of 26\,m
\cite{ludvigsen2014scientific,mogstad2022tautra}. The survey consists of parallel passes (survey lines) flown at alternating
headings of $\approx 52^\circ$ and $232^\circ$ with $\approx$50\,\% sidelap, covering 2.4\,km$^2$. Six
further passes cross the main lines in the north-west of the area (Figure~\ref{fig:survey}). This
configuration means that much of the seabed was imaged more than once. Each SAS image is a map of
backscattered acoustic intensity, that is, the strength of the echo.

\noindent\textbf{Cells and looks.}
We tile the survey with a fixed ground grid of $8.25\times8.25$\,m \emph{cells} (UTM~32N, axes rotated to
the line heading), corresponding to a SAS image of $224\times224$\,px at the native resolution of
$\approx$3.8\,cm per pixel. A \emph{look} is one full-aperture image of a cell from a pass, cut at the cell centre in the sensor
frame. Looks were
not rotated, resampled or interpolated. Two looks of a cell differ in bearing (median $91^\circ$, range
$0$--$180^\circ$), in range from the sonar to the cell centre (median $89$\,m, range $21$--$169$\,m, with
two looks of a cell differing by a median of $35$\,m) and in footprint (IoU median $0.84$, minimum
$0.55$). These differences are part of the look variation under consideration. Grid residuals are
$\le 0.034$\,m (pixel rounding), and looks of the same cell agree in position to a median of 2.0\,m where
image cross-correlation of the two looks finds a usable peak (S5).

This gives a dataset of 50{,}869 looks over 35{,}359 cells (Figure~\ref{fig:survey}). Of these, 12{,}686 cells
have two or more looks (10{,}069 with two, 2{,}410 with three, 207 with four), and every repeat-look
metric is computed on them. Single-look cells enter only the classification task. Example cells with their
looks are shown in Figure~S1.

\noindent\textbf{Coral labels.}
Labels are obtained from a coral-reef polygon map of the Tautra ridge produced from the 2012 survey
\cite{mogstad2022tautra,sture2018recognition}, which delineates coral (\textit{Desmophyllum pertusum})
reef. A cell is labelled positive if
more than 50\,\% of its area is covered by a coral polygon: 4{,}571 of all cells (13\,\%) and 1{,}861 of the
repeat-look cells (15\,\%).

\subsection{Pretrained encoders and preprocessing}\label{sec:reps}

\noindent\textbf{Natural-image encoder.}
DINOv3 ViT-S/16 \cite{simeoni2025dinov3}, a state-of-the-art self-supervised vision encoder, which sonar
studies have built on \cite{hamoda2026benthicdino} and compared against \cite{kwon2026minejepa}. Each sonar file (an along-track chunk of a line) is converted to log amplitude, clipped to its 0.1--99.9
percentiles and standardised per file and channel. Each look is then
replicated to three channels and encoded at $224\times224$\,px. The embedding is the mean of the 196
patch tokens after the final layer norm (384-D). Alternative readouts (CLS token, max- and GeM-pooling),
as commonly compared \cite{keetha2024anyloc}, were also assessed and did not affect
the interpretation of the results (S3).

\noindent\textbf{Sonar-pretrained encoder.}
BenthicDINO \cite{hamoda2026benthicdino}, a ConvNeXt-V2-T self-distilled on $\sim$1\,M side-scan sonar
(SSS) tiles with a view-invariance penalty against look geometry. As per the paper, preprocessing
(log amplitude, per-file 90th-percentile clip, fixed normalisation) was applied to each look, and the
readout fused the feature maps of the four network stages on a common grid and averaged over position
(768-D). Whilst trained on SSS imagery, it serves as a
useful additional encoder, as to our knowledge there is no publicly released self-supervised encoder
pretrained on SAS imagery.

\noindent\textbf{Autocorrelation function descriptor (ACF).}
As a non-learned reference we reproduce the ACF descriptor of \citet{cobb2011autocorrelation},
developed for SAS seabed segmentation. A Gaussian mixture fitted to each look's intensity
autocorrelation gives the paper's ten features plus the $K$-distribution shape, $D = 11$. The fit,
the preprocessing and the choices the paper leaves open are documented in S3.

\subsection{Embedding evaluation framework}\label{sec:eval}

\subsubsection{Standard single-look measures}\label{sec:eval-single}

These are computed on a single look per cell, exactly as they would be on an image dataset. Together
they describe the geometry of the embedding space: whether it is well formed, how many degrees of
freedom it uses, and whether it is clustered. Formulae and further details of the metric calculations
are given in S3, and Table~S1 lists the question each measure answers. Each representation is used at its native dimensionality, which does not affect
comparability between encoders, with the exception of the extrinsic and intrinsic dimensions (RankMe and
TwoNN respectively).

\noindent\textbf{M1 Effective dimensionality (RankMe).}
The entropy-based effective rank \cite{garrido2023rankme}, computed here on the $\ell_2$-normalised
embeddings. Low values indicate dimensional collapse, where a few directions dominate.

\noindent\textbf{M2 Intrinsic dimensionality (TwoNN).}
The TwoNN estimator \cite{facco2017twonn}, computed from the ratio of each cell's two nearest-neighbour
distances. Effective rank counts the linear directions of variance globally, whereas intrinsic dimension
(TwoNN) estimates the number of local degrees of freedom and is less sensitive to curvature: a curved sheet
has intrinsic dimension two however many linear directions it bends through. A low intrinsic dimension
alongside a high effective rank therefore indicates a curved, low-dimensional manifold spanning many
linear directions.

\noindent\textbf{M3 Cluster structure (cosine silhouette).}
We partition the embeddings with $k$-means at $k=5$ and report the mean silhouette under cosine
distance \cite{rousseeuw1987silhouettes}: for cell $i$ with mean distance $a_i$ to its own cluster and
$b_i$ to the nearest other cluster, $s_i = (b_i - a_i)/\max(a_i, b_i)$. Values near 1 indicate compact,
well-separated groups, values near 0 continuous or overlapping structure.

\noindent\textbf{M4 Cluster stability (ARI across seeds).}
We repeat $k$-means over ten random seeds and compute the adjusted Rand index (ARI) \cite{hubert1985comparing}
between the resulting partitions, the pair-counting agreement between two partitions corrected for
chance ($1$ = identical, $0$ = chance). High agreement indicates that the clusters reflect structure in
the data. Low agreement indicates arbitrary cuts through a continuum, in which case silhouette values
should not be read as evidence of discrete groups.

\noindent\textbf{M5 Downstream prediction: coral classification (AUC).}
A linear probe (class-balanced logistic regression, regularisation in S5) was trained on five spatial folds ($k$-means on cell coordinates) \cite{roberts2017crossvalidation} with training cells within
250\,m of the test fold removed, a buffer beyond the autocorrelation
range of the probe residuals ($\approx$100\,m; S6). Features were standardised on the
training fold, and then one look was chosen randomly per training cell and the area under the ROC
curve (AUC) computed on the test fold, repeated over 200 draws. Since a linear probe can predict well
along one fitted direction even when cells that are close in the embedding do not share the label, we
also ran a $k$-nearest-neighbour classifier on the same folds. It did not change the results and is
reported in S5.

\begin{table*}[t]
\centering\small
\caption{Single-look measures (M1--M5), each computed on one look per cell and repeated over 200 random
choices of look, and repeat-look reliability measures (M6--M10), each comparing the looks of the same
cell, for three representations. Repeat-look measures use the 12{,}686 repeat-look cells. Values are the
median [combined look-draw $\times$ 250\,m spatial block-bootstrap 95\,\% interval], best value per row
in bold (M4--M10). The probe uses five spatial folds with a 250\,m buffer. Chance
levels: ARI $0$, and
$\kappa = 0.31$ expected from two independent decisions at the probe's own accuracy (DINOv3; encoder-specific).}
\label{tab:main}
\begin{tabular}{rlccc}
\toprule
 & & DINOv3 & BenthicDINO & ACF \\
\midrule
\multicolumn{5}{l}{\emph{Single-look measures (one look per cell per draw)}} \\
 M1 & RankMe (effective rank) & 105 {\scriptsize\textcolor{gray}{[99, 109]}} & 116 {\scriptsize\textcolor{gray}{[107, 121]}} & 9.1 {\scriptsize\textcolor{gray}{[9.0, 9.2]}} \\
 M2 & TwoNN intrinsic dimension & 23.0 {\scriptsize\textcolor{gray}{[22.6, 23.4]}} & 13.9 {\scriptsize\textcolor{gray}{[13.6, 14.3]}} & 6.3 {\scriptsize\textcolor{gray}{[6.2, 6.4]}} \\
 M3 & silhouette, $k=5$ & 0.23 {\scriptsize\textcolor{gray}{[0.22, 0.25]}} & 0.26 {\scriptsize\textcolor{gray}{[0.25, 0.27]}} & 0.38 {\scriptsize\textcolor{gray}{[0.36, 0.39]}} \\
 M4 & ARI across seeds, $k=5$ & 0.996 {\scriptsize\textcolor{gray}{[0.99, 1.00]}} & \textbf{0.997} {\scriptsize\textcolor{gray}{[0.97, 1.00]}} & 0.997 {\scriptsize\textcolor{gray}{[0.98, 1.00]}} \\
 M5 & coral probe AUC & \textbf{0.93} {\scriptsize\textcolor{gray}{[0.84, 0.94]}} & 0.92 {\scriptsize\textcolor{gray}{[0.88, 0.93]}} & 0.78 {\scriptsize\textcolor{gray}{[0.71, 0.80]}} \\
\midrule
\multicolumn{5}{l}{\emph{Repeat-look reliability measures (looks of the same cell compared)}} \\
 M6 & dbICC & \textbf{0.20} {\scriptsize\textcolor{gray}{[0.15, 0.23]}} & 0.16 {\scriptsize\textcolor{gray}{[0.12, 0.19]}} & 0.13 {\scriptsize\textcolor{gray}{[0.10, 0.15]}} \\
 M7 & $r_{11}$ (Mantel) & 0.23 {\scriptsize\textcolor{gray}{[0.14, 0.34]}} & \textbf{0.28} {\scriptsize\textcolor{gray}{[0.21, 0.34]}} & 0.04 {\scriptsize\textcolor{gray}{[0.02, 0.06]}} \\
 M8 & IoK, $k=10$ & \textbf{0.013} {\scriptsize\textcolor{gray}{[0.011, 0.014]}} & 0.011 {\scriptsize\textcolor{gray}{[0.010, 0.014]}} & 0.001 {\scriptsize\textcolor{gray}{[0.001, 0.002]}} \\
 M9 & ARI across looks, $k=5$ & \textbf{0.08} {\scriptsize\textcolor{gray}{[0.05, 0.12]}} & 0.03 {\scriptsize\textcolor{gray}{[0.02, 0.04]}} & 0.020 {\scriptsize\textcolor{gray}{[0.010, 0.04]}} \\
 M10 & $\kappa$ of coral decision & \textbf{0.38} {\scriptsize\textcolor{gray}{[0.29, 0.47]}} & 0.33 {\scriptsize\textcolor{gray}{[0.23, 0.44]}} & 0.09 {\scriptsize\textcolor{gray}{[0.05, 0.12]}} \\
\bottomrule
\end{tabular}
\end{table*}

\subsubsection{Repeat-look reliability measures}\label{sec:relmeasures}

These measures answer different questions about the reliability (summarised in Table~S1). Encoder embeddings and the ACF
descriptor are $\ell_2$-normalised
and compared by cosine distance, as per the encoder evaluation itself \cite{simeoni2025dinov3}. All
distances below are squared Euclidean distances between unit vectors, $d^2_{ab} = \|x_a - x_b\|^2$ between
looks $a$ and $b$, which is twice the cosine distance.

\noindent\textbf{M6 Vector reliability (dbICC).}
We ask whether a cell's embedding is the same on an alternate look. This is test--retest reliability in the sense of classical measurement \cite{nunnally1978psychometric},
applied to a vector-valued measurement as a per-cell variant of the distance-based ICC of \citet{xu2021dbicc}. Let $L_i$ be the set of looks of
cell $i$ and $n$ the number of cells. Let $d^2_{\mathrm{within},i}$ be the mean squared distance
over the look pairs of cell $i$, and
\begin{equation}
\bar d^2_{\mathrm{between}} = \frac{1}{\sum_{i \ne j} |L_i||L_j|} \sum_{i \ne j} \sum_{a \in L_i}
\sum_{b \in L_j} d^2_{ab}
\label{eq:m2}
\end{equation}
the mean over all pairs of looks of different cells. Then
\begin{equation}
\rho_i = 1 - d^2_{\mathrm{within},i} / \bar d^2_{\mathrm{between}} ,
\qquad \mathrm{dbICC} = \frac{1}{n}\sum_i \rho_i .
\label{eq:m3}
\end{equation}
Intuitively, $\rho_i$ is how much closer a cell's looks are to each other than a random other cell:
$\rho_i \to 1$ as the looks coincide, $\rho_i = 0$ when an alternate look is no closer than a random
cell, and $\rho_i < 0$ when it is farther. The numerator, the mean distance between two images of the
same target, is the \emph{average pairwise cosine distance} with which \citet{thiringer2026scanner}
measured scanner sensitivity in pathology encoders. Dividing by the between-cell distance removes the
collapse confound they note. The classical one-way ICC on all looks and the rank-based discriminability
of \citet{bridgeford2021discriminability} give the same picture and are reported in S5.

\noindent\textbf{M7 Geometry reliability ($r_{11}$).}
This asks whether the distances between cells are the same on an alternate look. Many uses of an
embedding space use the pattern of distances between targets, e.g.\ retrieval,
clustering, site selection and representational similarity analysis \cite{kriegeskorte2008rsa}, which
depend on geometry reliability rather than on vector reliability. For each cell we draw one look $a_i$
and an alternate look $b_i$ at random. The two choices give two sets of pairwise distances between the
repeat-look cells, $A_{ij} = d^2_{a_i a_j}$ and $B_{ij} = d^2_{b_i b_j}$, and $r_{11}$ is their Pearson
correlation over the $\binom{n}{2}$ cell pairs $i<j$ (the \emph{Mantel correlation} of
\citet{thiringer2026scanner}). This is the test--retest reliability of a representational geometry as
estimated in representational analysis \cite{walther2016reliability}, and it bounds how
well the geometry can agree with anything else \cite{spearman1904proof}. Intuitively, $r_{11}$ indicates
whether the global pattern of distances repeats. A representation can have low vector reliability
(dbICC) and high geometry reliability ($r_{11}$) if, for instance, every cell shifted together.

\noindent\textbf{M8 Neighbourhood overlap (IoK).}
IoK asks whether the \emph{local} neighbourhood is reliable. For each look, the fraction of its $k=10$ nearest neighbours (among looks of other cells) that are also
among the $k$ nearest neighbours of the cell's alternate look, adapted from the intersection-over-$k$ of
\citet{thiringer2026scanner}, chance $k/N$.

\noindent\textbf{M9 Partition stability across looks (ARI).}
The adjusted Rand index (ARI) between $k$-means partitions ($k=5$) computed from different looks of
the same cells, read against M4: if the seed-to-seed ARI is near 1 and the look-to-look ARI near 0, the
clusters are stable to the algorithm and unstable to the measurement.

\noindent\textbf{M10 Prediction stability across looks ($\kappa$).}
We ask whether the classifier predicts the same for cells on an alternate look. This is different from
M5, since the AUC is a population number and says nothing about any single cell. We use Cohen's
$\kappa$ \cite{cohen1960coefficient} on the probe's test-set predictions for the looks of a cell, the two-look analogue of the
Fleiss' $\kappa$ of \citet{thiringer2026scanner}.
$\kappa = 1$ means every cell gets the same decision on both looks. The floor is the $\kappa$ expected
from two independent decisions at the probe's own accuracy, which we report alongside. The decision
threshold is set within each training fold so that the fraction of training looks called positive
equals the fraction that are positive.

This list is not exhaustive of the metrics that could be used. For instance, we do not report
trustworthiness and continuity \cite{venna2005local}, as they compare an embedding with a reference
space, and two looks are two equally noisy measurements with no reference.

\subsubsection{Variance components (generalizability theory)}\label{sec:gtheory}

As noted in Section~\ref{sec:related}, the coefficient depends on what variation is treated as error.
Generalizability theory \cite[ch.~2, pp.~30--35]{brennan2001generalizability} makes this explicit. We model the embedding, $x$,
of look $l$ of cell $c$ as
\begin{equation}
x_{cl} = \mu_c + \nu_g + \nu_{c\times g} + \nu_e ,
\label{eq:gmodel}
\end{equation}
where $\mu_c$ is the cell's look-independent embedding (the \emph{universe score}), defined as the mean
over all look geometries it could be imaged from. $\nu_g$ is the effect of the look geometry $g$ shared by
all cells. $\nu_{c\times g}$ is the cell-specific effect of that geometry, and $\nu_e$ collects residual effects,
e.g.\ speckle and residual motion-compensation error. Because no cell is seen twice under the same geometry in this
dataset, $\nu_{c\times g}$ and $\nu_e$ cannot be separated and are estimated together as one residual,
$\nu_{c\times g,e}$. Furthermore, $\nu_g$ can be broken down along any set of independent axes that fix the
sonar's position relative to the cell. We use range ($\nu_r$), altitude ($\nu_a$) and bearing
($\nu_b$), with bearing the direction from the cell to the sonar and range the distance between them.
Altitude was near-constant (26\,m), so $\nu_a = 0$ and
\begin{equation}
\nu_g = \nu_b + \nu_r .
\label{eq:gsplit}
\end{equation}

We report two reliability coefficients from this model, values between 0 and 1, and the variance
components they are built from.
\begin{enumerate}\setlength\itemsep{2pt}
\item The \emph{index of dependability} \cite[eq.~2.41]{brennan2001generalizability}: all variation from
$\mu_c$ counts as error,
\begin{equation}
\Phi = \frac{\sigma^2(c)}{\sigma^2(c) + \sigma^2(b) + \sigma^2(r) + \sigma^2(c\times g,e)} ,
\label{eq:phi}
\end{equation}
with $\sigma^2(\cdot)$ the variance of each component of variation. $\Phi$ is the variance-components
counterpart of dbICC (M6).
\item The \emph{generalizability coefficient} \cite[eq.~2.40]{brennan2001generalizability}: the reliability
once the variation from $\nu_b$ and $\nu_r$ has been removed,
\begin{equation}
E\rho^2 = \frac{\sigma^2(c)}{\sigma^2(c) + \sigma^2(c\times g,e)} ,
\label{eq:erho}
\end{equation}
This is an adjusted intraclass correlation in the sense of \citet{nakagawa2017r2}. We compute
this for two reasons: to estimate the reliability the encoder would have if these effects were accounted for, and to
demonstrate the method.
\item The \emph{variance components} $\sigma^2(c)$, $\sigma^2(b)$, $\sigma^2(r)$ and $\sigma^2(c\times g,e)$
themselves.
\end{enumerate}

The effects are fitted by least squares on within-cell contrasts, cross-fitted over two splits of the
cells, and subtracted. The components are then estimated per embedding dimension and summed over
dimensions (S4).

\subsubsection{Reliability of the probe direction}\label{sec:probedir}

The repeat-look measures above (M6--M9) treat all dimensions of the embedding alike, so to examine how
reliability bears on the linear-probe result we measure the reliability of the probe direction. The
probe's weight vector $w$ maps each look's embedding, $x$, to the scalar score $s = w^\top x$. Its reliability
across looks is dbICC (M6), \eqref{eq:m3}, applied to that scalar, which for a scalar is the classical
intraclass correlation (ICC). To avoid
test leakage, $w$ is fitted on training cells and the ICC computed on held-out cells (a two-fold 500\,m
checkerboard split with a 100\,m buffer, the residual autocorrelation range, swapped and averaged).

\subsection{Uncertainty and reporting}\label{sec:reporting}

Every statistic is reported as the median over 200 random draws of one look per cell, with a 95\,\%
interval that combines the look draws with a 250\,m spatial block bootstrap \cite{lahiri2003resampling} of the cells (63 blocks; S6). These blocks are used only for the intervals; test cells are held out separately, by the probe's five spatial folds and the
probe-direction checkerboard.

%% file: sec_results.tex
\section{Results}\label{sec:results}

\begin{figure}[t]
\centering
\includegraphics[width=\linewidth]{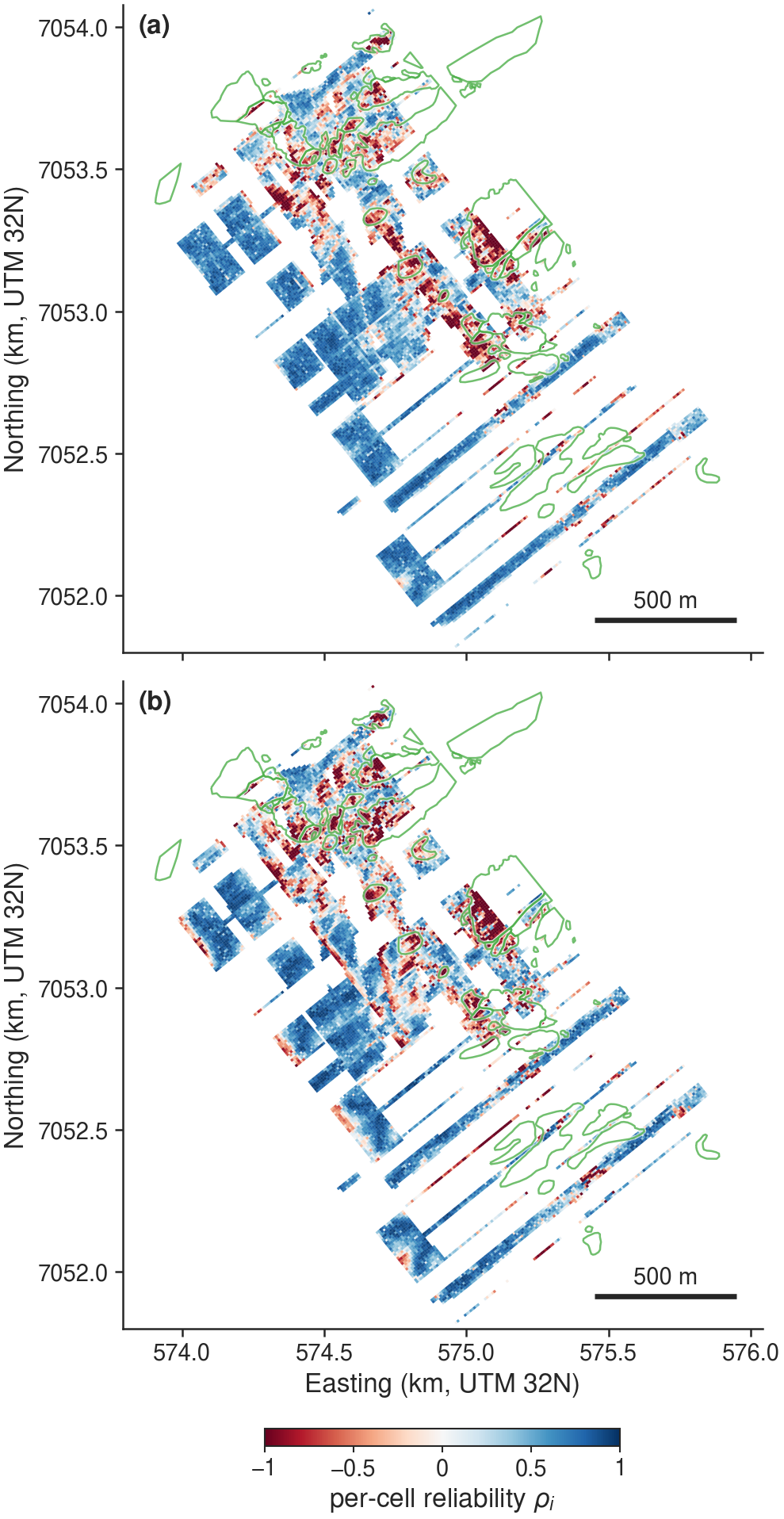}
\caption{Per-cell vector reliability $\rho_i$ for (a) DINOv3 and (b) BenthicDINO on the 12{,}686 cells
with two or more looks (blue: high, red: negative; green outlines: 2012 coral-reef polygons).
Statistics in Section~\ref{sec:results-repeat}.}
\label{fig:maps}
\end{figure}

\subsection{Standard single-look measures}\label{sec:results-single}

Table~\ref{tab:main} (top) gives the standard quality metrics and the coral probe. They are all stable to
the choice of look: across 200 random draws the quality metrics move by at most $2$\,\%, the probe AUC by
$\pm 0.003$ (S5), and $k$-means partitions recomputed with different seeds agree. The two encoders decode
coral well and the texture descriptor less so (AUC $0.93$, $0.92$ and $0.78$; Table~\ref{tab:main}). Nothing in this table distinguishes a reliable
representation from one that is not.

\subsection{Repeat-look reliability}\label{sec:results-repeat}

\paragraph{Vector reliability (M6).}
Across the repeat-look measures (Table~\ref{tab:main}, bottom), the sonar pretraining and view-invariance
penalty of BenthicDINO did not make its representation more reliable across looks than DINOv3's. For DINOv3, dbICC is $0.20$: an alternate look of a cell lands, on average, 80\,\% as far
from the first as a random other cell. The per-cell values are skewed, with median $\rho_i$ $0.35$ and a
long negative tail from the reef (histogram in S5). BenthicDINO is lower ($0.16$) and ACF lower still
($0.13$).

Figure~\ref{fig:maps} maps $\rho_i$. The two encoders agree on which cells are unreliable (Spearman
$0.67$ between maps). $\rho_i$ is negative on the reef: cells with reef cover (more than 50\,\% coral polygon) have median
$\rho_i$ of $-0.32$ (DINOv3) and $-0.39$ (BenthicDINO) measured against the whole survey, cells without
$0.44$ and $0.39$. Measured
against other reef cells (S5) it is close to zero ($+0.06$ [$0.00$, $0.13$]): two looks of a reef cell are barely
more alike than two different reef cells, while cells without any coral polygon remain distinguishable
from each other ($0.36$ against other such cells). The nearest-neighbour galleries (Figure~S3) exhibit how this unreliability
manifests, with examples of coral and non-coral cells, the reef cell presenting mounds from one
direction and near-featureless texture from another. The query look (red) is clearly similar to its
nearest neighbours in embedding space, so the encoder is grouping well. However, the alternate look of
the same cell (blue) can be very far away in embedding space.

\paragraph{Geometry reliability (M7).}
$r_{11}$ is $0.23$ for DINOv3, $0.28$ for BenthicDINO and $0.04$ for ACF. The relative
geometry does not repeat across looks, so the neighbourhood and partition measures (M8, M9) are expected
to be poor.

\paragraph{Neighbourhoods, partitions and decisions (M8--M10).}
IoK is $0.013$ (chance $0.001$). $k$-means partitions built from different looks of the same cells
largely do not agree, ARI $0.08$, against $0.996$ when only the seed changes.
The coral decision of the probe (M10) agrees between two looks of a cell $87$\,\% of the time,
$\kappa=0.38$ [0.29, 0.47], against the floor of $\kappa=0.31$ [0.21, 0.40].
Cross-pass retrieval fails as well: a look's same-cell partner was among its ten nearest looks for $0.32$\,\% of looks
(DINOv3; chance $0.02$\,\%; self-recall@10, S5).

\subsection{Generalizability theory: variance components}\label{sec:results-where}

\begin{figure}[t]
\centering
\includegraphics[width=0.9\linewidth]{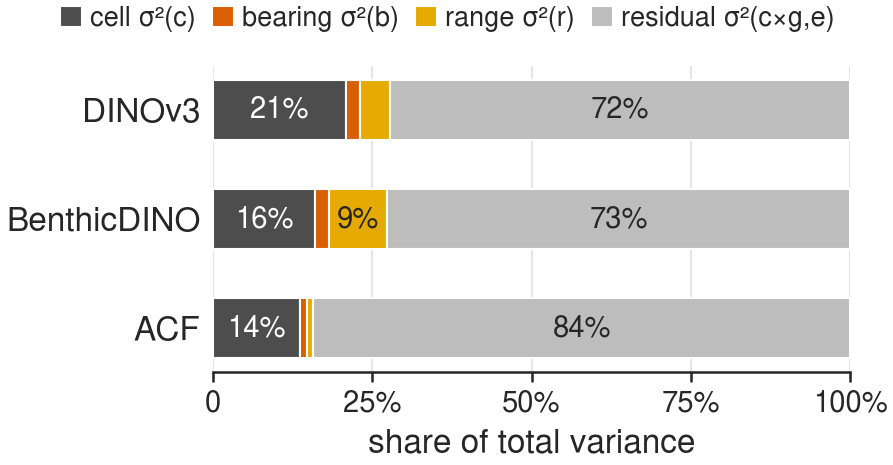}
\caption{Variance components of the embeddings (Section~\ref{sec:gtheory}) as shares of total variance: cell
$\sigma^2(c)$, the survey-wide bearing $\sigma^2(b)$ and range $\sigma^2(r)$ effects, and the residual
$\sigma^2(c\times g,e)$, which is the look-to-look variation not shared across cells (cell-specific geometry
effects and noise). Leaving the cell component out, the within-cell variance splits into bearing 3\,\%
(DINOv3) / 3\,\% (BenthicDINO), range 6\,\% / 11\,\% and residual 91\,\% / 87\,\% (ACF 1\,\%, 1\,\%,
98\,\%).}
\label{fig:components}
\end{figure}

Figure~\ref{fig:components} gives the variance components. For DINOv3 the cell accounts for 21\,\% of the
variance, bearing for 2\,\%, range for 5\,\% and the residual $\nu_{c\times g,e}$ for 72\,\%. From these
components $\Phi = 0.21$, the variance-components counterpart of dbICC ($0.20$, which weights cells
equally). $E\rho^2$ for the $\nu_g$-corrected embeddings is $0.22$ [0.18, 0.27], a marginal improvement. BenthicDINO ($\Phi$ $0.16 \to 0.18$) and ACF ($0.14 \to 0.14$) behave the same way. For ACF the
residual is 84\,\% of the total variance and 98\,\% of the within-cell variance.

\subsection{Reliability of the probe direction}\label{sec:results-direction}

The ICC of the coral probe score $w^\top x$ across looks is $0.60$ [0.44, 0.66] for DINOv3 (splits and floors in S5). Random
directions give a median ICC of $0.18$ (95th percentile $0.32$), the ten leading principal components at
most $0.38$, and probes fitted to permuted labels at most $0.06$. The direction the probe reads is about
3.0$\times$ as reliable across looks as the embedding as a whole (dbICC $0.20$), and more reliable than
any of the unsupervised directions we tested. Within a class the same score is far less reliable, with
ICC $0.38$ [0.26, 0.45] among non-reef cells and $0.30$ [0.13, 0.41] among reef cells. The direction
is reliable for the habitat type the probe was fitted to rather than for individual cells: within a class
it falls to about the reliability of the full embedding off the reef ($0.38$ against $0.36$), and stays low on
the reef ($0.30$), though above the full embedding there ($+0.06$).

%% file: sec_discussion.tex
\section{Discussion}\label{sec:discussion}

The quality of the encoders on SAS data is not under consideration per se. We discuss the key points
regarding the reliability of their embeddings.

\noindent\textbf{Standard evaluation is look-insensitive.} Standard, single-look evaluation (M1--M5) does
not reveal look-to-look variation. This hides variation that matters when the same physical
space is imaged more than once, as in repeated acoustic surveys \cite{roche2018control,lurton2018azimuth}. Assessing
reliability in these cases needs other measures and methods.

\noindent\textbf{Reliability is cell dependent.} With this in mind, the paper examined a suite of repeat-look
measures (M6--M10). Vector reliability (M6, dbICC) varies strongly between cells. On the reef there is little
cell-level information in the embedding beyond the reef itself, for every representation. Because reef
cells are not identifiable across looks (Section~\ref{sec:results-repeat}), any use that needs patch
identification, such as retrieval or change detection between surveys, will fail. Cross-pass re-identification fails outright on the survey as a whole, as self-recall (S5) and
partition stability (M9) show. Off the reef the embeddings are moderately
reliable, with an alternate look landing noticeably closer than a random cell. Reef pairs viewed
from the same bearing have reliability close to zero, while reef pairs viewed from different bearings are
strongly negative, which points at anisotropy, and the reef deficit persists within each bearing stratum (S5). Geometry
reliability (M7, $r_{11}$) is also low, which bears on retrieval, clustering and site selection (M8, M9).

\noindent\textbf{The AUC remains high.} Despite this, the coral AUC (M5) remains high, in large part
because the probe's weight vector finds a reliable direction. The looks of a reef cell can therefore
presumably be far apart in the embedding and still fall on the same side of the probe's hyperplane. The
AUC is insensitive to the choice of look because it ranks the whole population, and exchangeable looks
only reshuffle scores within each class. This breaks down at the per-cell level, which a repeat-look
measure (M10, $\kappa$) shows. Decision agreement across looks is slightly above what two independent predictions
would give, but falls in the range conventionally read as fair at best \cite{landis1977measurement}.

\noindent\textbf{SSS encoder was not more reliable.} BenthicDINO, pretrained on side-scan imagery with a
view-invariance penalty, was no more reliable across looks than DINOv3 and failed in the same places
(Figure~\ref{fig:maps}b; galleries in Figure~S4). A likely reason is that real-aperture side-scan is far removed from SAS. Along-track
resolution degrades with range, and speckle and shadow statistics differ \cite{lurton2010introduction}. Further, DINOv3 may benefit
from the enormous amounts of data it was trained on, despite the domain mismatch. This suggests that
SAS-specific models are needed, and at scale.

\noindent\textbf{The generalizability framing.} The bearing effect ($\nu_b$) is small, consistent with the
argument that over a representative sample of seabed there is no survey-wide bearing effect \cite{lurton2018azimuth}. In practice
surveys will likely have a non-zero effect due to persistent seabed characteristics, in this case the
Tautra ridge slope \cite{ludvigsen2014scientific} and aligned ripple fields.

The range effect ($\nu_r$) and altitude ($\nu_a$) are
properties of the sonar instrument and will act survey-wide, as range affects grazing angle,
backscatter level, shadow length, across-track resolution and attenuation \cite{lurton2010introduction}. Future studies
should account for this in the survey design to obtain a better account of reliability.

The decomposition separates the survey-wide effects from the rest, allowing an estimation of what
preprocessing, augmentation or invariance training could gain. Which of these corrections is best
remains an open question, e.g.\ incidence-angle modelling within the classifier in
satellite radar \cite{lohse2020seaice}, statistical harmonisation of site effects in neuroimaging
\cite{pomponio2020harmonization}, augmentation or natural positives in the loss in representation
learning \cite{yamada2022geoclr,manas2021seasonal}.

Another viable reliability framing exists in which geometry is part of the signal, that is, the
look-geometry facet is fixed, so that $\nu_g$ and $\nu_{c\times g}$ are part of the signal \cite[\S3.3]{webb2006gtheory}. Then the representation is the cell seen from a given geometry,
$\mu_c + \nu_g + \nu_{c\times g}$, and only $\nu_e$ is error. Since sonar backscatter is anisotropic with
respect to look direction \cite{lurton2018azimuth}, $\nu_{c\times g}$ holds real structure
and this representation is the physically natural one, although this might not be necessary for all
tasks. Its coefficient needs $\sigma^2(c\times g)$ separated from $\sigma^2(e)$, which requires repeat looks at
matched look geometry. The lawnmower survey used here does not provide them.

\noindent\textbf{Limitations.} The study uses one scanner and one
survey. A natural extension is to data
from different scanners, settings and resolutions, the sonar counterpart of the scanner shifts studied
elsewhere \cite{thiringer2026scanner}. The same decomposition takes scanner and settings as further facets. The
study also compares only three representations on one simple downstream task with a single label set.
More tasks would sharpen the analysis.

\noindent\textbf{Implications.} Whilst one may question the value of this given the good downstream performance, the analysis is
useful in the following ways.

(a) It quantifies the look-to-look variation of a learned representation of a re-measured target and
attributes it to named sources. We are not aware of this having been done for seabed or remote-sensing
imagery, and the method applies to any sensor that images the same target more than once.

(b) The results from (a) can be used to assess architectural and training decisions. The generalizability study
identifies which variations to target and the repeat-look metrics show whether a change helped. It should be
noted that the standard way to handle these variations is through augmentations, which are functions of
the image. So augmentations could be used to simulate the desired variances, but the test can only be
made on repeat looks. Better than this are methods that incorporate natural positives into the loss
function, such as GeoCLR \cite{yamada2022geoclr}, Tile2Vec \cite{jean2019tile2vec} and Seasonal
Contrast \cite{manas2021seasonal}.

(c) A decision study (D-study) is the natural follow-up to a generalizability study
\cite{brennan2001generalizability,webb2006gtheory}. It uses the estimated variance components to project
the reliability of alternative survey designs before they are flown, such as the number of looks per
cell and the look geometry needed to reach a required reliability. The per-cell reliability maps (M6, dbICC)
can likewise be used to propose new survey sites.

Given this, we propose that repeat-look reliability, which is free wherever a survey has sidelap, be
part of the tool-kit for repeat-look data, and that surveys intended to characterise a representation
include the key sources of variation, for example cells at matched look geometry.

%% file: sec_supp.tex
\section{Code and data}
The SAS imagery and the coral map are held by the survey owners and are available from them on request, as for the earlier studies of this survey. Requests for the derived look embeddings and the analysis code can be directed to the authors.

\section{Dataset}
\begin{figure*}[t]\centering
\IfFileExists{figures/survey/look_pair_contact_sheet.jpg}{\includegraphics[width=\textwidth]{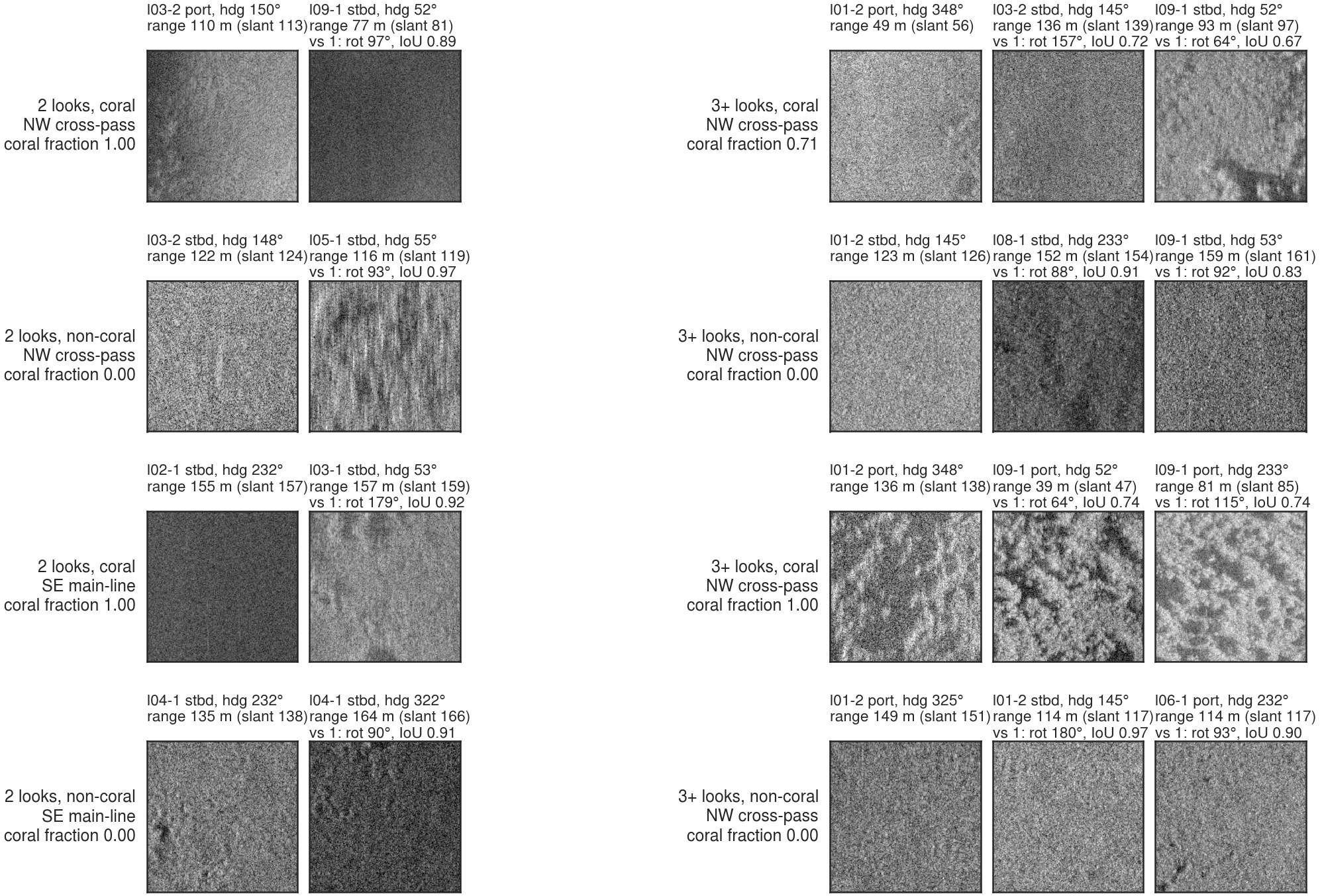}}{}
\caption{Example cells with all their looks (Section~3.1 of the main text). Eight cells chosen a priori without reference to any embedding: four with two looks and four with three or more, alternating coral and non-coral, from the north-west crossing region and the south-east main lines. Each row of the two blocks shows every look of one cell as cut, in its native orientation (rows along the pass heading, port not mirrored), annotated with pass (a pass can include legs flown on different headings), channel, heading, ground and slant range, and, for looks after the first, the rotation and footprint IoU against the first look.}
\label{fig:s-contact}\end{figure*}

\section{Representations}
\paragraph{ACF descriptor (Section~3.2).} A four-component Gaussian mixture (one paired non-zero-mean component and two zero-mean components) is fitted by truncated-data EM to the normalised intensity autocorrelation of each look over lags up to 56\,px, giving the paper's ten features (the correlation lengths $l_x$, $l_y$ and weights $\alpha$ of the three components and the ripple spacing $|\mu_1|$) plus the $K$-distribution shape $\hat\nu$. The fit is run on the whole 224-px look, converted back to intensity, without sub-windows or further normalisation. The features are z-scored over all looks, as in the paper, before $\ell_2$ normalisation and cosine distance, and standardised on the training fold for the probe. The $\hat\nu$ estimator is the single-look log-moment estimator, the point-spread function is a single global fit to the speckle peak, and the EM is initialised from the spectral peak, with eigenvalue clamps.
\paragraph{Readout and distance (Section~3.2).} Reading the CLS token, max- or GeM-pooling instead of the patch-token mean gives dbICC $0.205$, $0.198$ and $0.187$ (patch-token mean $0.196$), $r_{11}$ $0.27$, $0.22$ and $0.23$ ($0.23$). Every readout is within $0.04$ of the patch-token mean on these rows, and the reef deficit persists under every readout. The readouts give identical kNN overlap and coral AUC within $0.008$; only the shape statistics move (RankMe 100--134, TwoNN 23--28). The reef cells are the least reliable under every readout (Spearman $\rho_i$ vs reef fraction $-0.42$ to $-0.46$). Cosine, Euclidean and centred distances give dbICC $0.196$--$0.217$ on the same pairs. Every difference stated above holds at 100, 250 and 500\,m blocks (S6).
\paragraph{Single-look metric definitions (Section~3.3.1).} Let $Z \in \mathbb{R}^{N \times D}$ hold one $\ell_2$-normalised embedding per cell. \emph{RankMe} (M1): with $\sigma_1 \ge \dots \ge \sigma_D$ the singular values of $Z$ and $p_k = \sigma_k / \sum_j \sigma_j$, $\mathrm{RankMe}(Z) = \exp(-\sum_{k=1}^{D} p_k \log p_k)$ \cite{garrido2023rankme}; it is a label-free indicator of collapse, comparable across checkpoints of one model rather than across architectures, and is reported descriptively. \emph{TwoNN} (M2): with $r_1(i) \le r_2(i)$ the distances from cell $i$ to its two nearest neighbours and $\mu_i = r_2(i)/r_1(i)$, the intrinsic dimension $d$ is the slope of $-\log(1-F(\mu))$ against $\log \mu$, where $F$ is the empirical distribution of $\mu$ \cite{facco2017twonn}. \emph{Silhouette} (M3): for cell $i$ with mean cosine distance $a_i$ to its own $k$-means cluster and $b_i$ to the nearest other cluster, $s_i = (b_i - a_i)/\max(a_i, b_i)$ \cite{rousseeuw1987silhouettes}. \emph{ARI} (M4, M9): the pair-counting agreement between two partitions corrected for chance \cite{hubert1985comparing}.

\begin{table*}[t]\centering\small
\caption{The measures used in the paper and the question each answers (Section~3.3). M1--M5 use one look per cell; the rest use two or more looks of the same cell. ``Bears on'' names uses of an embedding that depend on the measure.}
\label{tab:s-metrics}
\begin{tabular}{@{}p{0.20\textwidth}p{0.43\textwidth}p{0.31\textwidth}@{}}
\toprule
Measure & Question it answers & Bears on \\
\midrule
\multicolumn{3}{@{}l}{\emph{Single look per cell}} \\
M1 RankMe & How many directions of the embedding space are used? & capacity; collapse \\
M2 TwoNN & How many degrees of freedom do the embeddings have locally? & capacity \\
M3 Silhouette & Do the embeddings form compact, separated clusters? & clustering \\
M4 ARI across seeds & Are those clusters stable to the clustering algorithm's randomness? & clustering \\
M5 Coral probe AUC & Does one look per cell support coral classification? & habitat classification \\
\midrule
\multicolumn{3}{@{}l}{\emph{Repeat looks of the same cell}} \\
M6 dbICC, per-cell $\rho_i$ & Is a cell's embedding the same on an alternate look, relative to the distance between cells? & any per-cell use: re-identification, change detection \\
M7 $r_{11}$ & Are the distances between cells the same on an alternate look? & retrieval, clustering, site selection, RSA; bounds agreement with other data \\
M8 IoK & Are a look's nearest neighbours the same cells on the alternate look? & nearest-neighbour search \\
M9 ARI across looks & Do clusters built from different looks of the same cells agree? & habitat partitions and maps \\
M10 $\kappa$ & Does the classifier give a cell the same decision on an alternate look? & per-cell maps from a classifier \\
Self-recall@10 (S5) & Is a look's same-cell partner among its ten nearest looks? & cross-pass retrieval \\
ICC(1) (S5) & The M6 question, answered with the classical one-way ANOVA estimator & check on M6 \\
Discriminability (S5) & The M6 question, answered from the ranks of distances only (unchanged by any monotone change of distance) & check on M6 \\
\midrule
\multicolumn{3}{@{}l}{\emph{Variance components and probe direction}} \\
$\Phi$ & How much of the embedding's variance is the cell, counting every look effect as error? & counterpart of M6 \\
$E\rho^2$ & How reliable would the embedding be once bearing and range effects are removed? & what correcting look geometry could gain \\
Variance components $\sigma^2(\cdot)$ & Where does the variation come from: cell $\sigma^2(c)$, bearing $\sigma^2(b)$, range $\sigma^2(r)$ or residual $\sigma^2(c{\times}g,e)$? & which correction to attempt \\
Probe-direction ICC & Is the classifier's score reliable across looks, even when the whole embedding is not? & why classification works despite low M6 \\
\bottomrule
\end{tabular}
\end{table*}

\section{Estimation of the variance components (Section~3.3.3)}
The facet effects and components (main-text Section~3.3.3) were estimated by: (i) choosing the levels of each facet, in this case four $45^\circ$ sectors for
bearing (the looks fall in four separated clusters) and six equal-count bins for range (a continuous covariate, or a Gaussian-process term
in the geometry, can take the place of levels in the fit below); (ii) fitting $\nu_b$ and $\nu_r$ by least squares on within-cell
contrasts, so that $\mu_c$ cannot enter the estimates,
\begin{equation}
\begin{split}
\hat\nu_b, \hat\nu_r = \arg\min \sum_c \sum_{l \in c}
\big\lVert (x_{cl} - \bar x_c) \qquad\qquad\qquad \\
- (\nu_{b(l)} - \bar\nu_{b,c}) - (\nu_{r(l)} - \bar\nu_{r,c}) \big\rVert^2 ,
\end{split}
\label{eq:fit}
\end{equation}
where $\bar x_c$, $\bar\nu_{b,c}$ and $\bar\nu_{r,c}$ are means over the looks of cell $c$; the fit was
cross-fitted over two splits of the cells to avoid leakage; (iii) obtaining $\sigma^2(b)$ and
$\sigma^2(r)$ as the look-weighted variances of the fitted effects,
\begin{equation}
\begin{aligned}
\sigma^2(b) &= \textstyle\sum_{j} \max\big(0,\ \tfrac{1}{N} \sum_{l} \hat\nu_{b(l),j}^2 - \hat s^2_{b,j}\big) ,\\
\sigma^2(r) &= \textstyle\sum_{j} \max\big(0,\ \tfrac{1}{N} \sum_{l} \hat\nu_{r(l),j}^2 - \hat s^2_{r,j}\big) ,
\end{aligned}
\label{eq:facetvar}
\end{equation}
with $N$ the number of looks, $j$ the embedding dimension, the effects centred to zero mean over looks and $\hat s^2_{\cdot,j}$ the
least-squares sampling variance of the fitted effects; these are variances of fitted effects less their sampling variance, not
expected-mean-square estimates; (iv) subtracting the
fitted effects from each embedding to obtain the corrected embedding
\begin{equation}
x'_{cl} = x_{cl} - \hat\nu_{b(l)} - \hat\nu_{r(l)} ,
\label{eq:corrected}
\end{equation}
and obtaining $\sigma^2(c)$ and the within-cell variance by the between-cell versus within-cell
comparison of dbICC (M6), using Henderson's method-of-moments estimator for unequal group sizes
\cite[\S7.1.2]{brennan2001generalizability}, on the corrected embeddings $x'$; $\Phi$ (main-text Eq.~5) adds the fitted
$\sigma^2(b)$ and $\sigma^2(r)$ back to its denominator and $E\rho^2$ (main-text Eq.~6) does not. Applied to the raw embeddings $x$
without geometry effects, the same estimator gives the one-way coefficient ($0.197$ for DINOv3), the counterpart of dbICC. All components are computed per embedding
dimension and summed over dimensions \cite{shou2013i2c2,xu2021dbicc}; negative per-dimension estimates of $\sigma^2(c)$ and the
within-cell variance are retained so that the sums stay unbiased, while the facet variances are floored at zero per dimension.

\paragraph{Estimator check.} A one-step linear mixed model fitted by REML \cite{patterson1971recovery} per embedding dimension (bearing sector and range bin as fixed effects, random cell intercept) gives $E\rho^2 = 0.218$ for DINOv3 against $0.224$ for the two-stage estimate (per-dimension $\sigma^2(c)$ correlation $1.000$). The one-step value is slightly lower because the mixed model's fixed effects also use between-cell contrasts, through which habitat confounded with look geometry is attributed to geometry; restricting the one-step fit to within-cell contrasts gives $0.241$, and the two-stage value lies between the two. All three lie inside the 250\,m block-bootstrap interval [0.18, 0.27].

\section{Additional results}
\begin{figure}[h]\centering\includegraphics[width=0.75\linewidth]{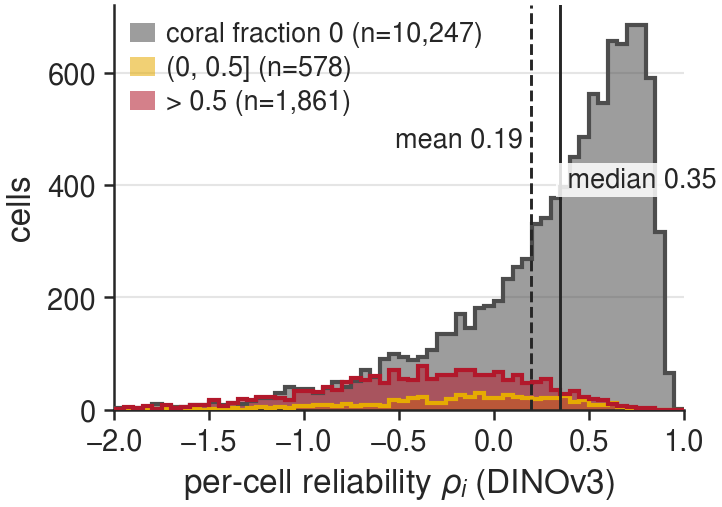}
\caption{Per-cell reliability $\rho_i$ (DINOv3) by coral fraction; dashed line mean, solid line median.}\label{fig:s-rhohist}\end{figure}

\paragraph{One-way ICC and discriminability (Section~3.3.2, M6).} The classical one-way intraclass correlation on all looks of the repeat-look cells is $0.20$ [0.15, 0.23] for DINOv3, $0.15$ [0.11, 0.19] for BenthicDINO and $0.13$ [0.10, 0.15] for ACF, and the rank-based discriminability of \citet{bridgeford2021discriminability} is $0.64$ [0.60, 0.69], $0.59$ [0.55, 0.63] and $0.59$ [0.57, 0.60] (chance $0.5$), the same ordering as dbICC.
\paragraph{Distribution of per-cell reliability (Section~4.2).} Figure~\ref{fig:s-rhohist} shows $\rho_i$ for DINOv3 by coral fraction. The distribution is skewed: mean $0.19$ (dbICC $0.20$), median $0.35$, 30\,\% of cells negative. Cells with coral fraction 0 have median $0.46$ (21\,\% negative), cells with fraction in (0, 0.5] median $-0.08$ (58\,\% negative), and reef cells (fraction $>0.5$) median $-0.32$ (72\,\% negative). The mean is pulled below the median by the reef tail.
\paragraph{Stratified between-cell denominator (Section~4.2).} The between-cell distance in $\rho_i$ is a survey-wide mean, 65\,\% of whose pairs are sediment--sediment, where \emph{sediment} here and in the next paragraph means coral fraction 0 (a subset of the main text's non-reef class, which also includes the cells with fraction in (0, 0.5]). By class pair (DINOv3, cosine): reef--reef $0.214$, sediment--sediment $0.127$, reef--sediment $0.205$, global $0.151$. With each cell's class-matched denominator, median $\rho_i$ is $+0.06$ [$0.00$, $0.13$] on the reef (global $-0.32$) and $0.36$ on sediment (global $0.46$), 44\,\% of reef cells are negative (global 72\,\%), dbICC is $0.16$ (global $0.20$), and the two maps correlate at Spearman $0.96$. BenthicDINO behaves the same ($-0.39 \to +0.10$); ACF is near zero under both.
\paragraph{Reef reliability by bearing stratum (Section~5).} For a stratified sample of 1{,}500 repeat-look cells (coral fraction 0, (0, 0.5] and $>0.5$), the two looks of each cell were cross-correlated after a 2\,m high-pass, with a usable peak defined against a displaced-window null. Usable pairwise offsets have median 2.0\,m (interquartile range 1.0--3.0\,m); pass-pair consensus offsets have median 2.0\,m. A usable peak exists for 35\,\% of sediment pairs and 13\,\% of reef pairs viewed from different bearings, and for 48\,\% and 26\,\% when viewed from the same bearing. Re-cutting the second look at the offset of its cross-correlation peak and re-embedding changes reef pair reliability by $0.00$ to $+0.01$, the same as a placebo re-cut of equal size in a random direction. Within same-bearing pairs, median pair reliability is $-0.03$ on the reef and $0.39$ on sediment; within cross-bearing pairs, $-0.31$ and $0.47$. The reef deficit therefore persists within each bearing stratum and is not a consequence of reef cells being seen more often across bearings.
\paragraph{Self-recall@10 (Section~5).} Whether a look's same-cell partner ranks among the ten nearest of all 50{,}868 other looks under cosine distance (chance $0.02$\,\%); the cross-scanner match rate of \citet{thiringer2026scanner} at $k=10$. Where kNN overlap (M8) asks whether the neighbours of two looks are the same other cells, self-recall asks whether the alternate look is itself found. DINOv3 $0.32$\,\% [0.18, 0.48], BenthicDINO $0.11$\,\% [0.05, 0.21], ACF $0.10$\,\% [0.05, 0.16]; median rank of the partner 13{,}832 (DINOv3). Readouts (CLS, max, GeM) give identical values. Every representation finds the partner for under 0.5\,\% of looks.
\paragraph{Probe regularisation (Section~3.3.1, M5).} The regularisation strength $C$ (inverse $\ell_2$ penalty) was chosen for each representation once, on an inner split of the first training fold, and reused ($C=0.01$ for all three representations). AUC is flat in $C$ for DINOv3 and ACF, and the chosen value is the best on the grid for BenthicDINO.
\paragraph{$k$-nearest-neighbour classifier (Section~3.3.1, M5).} The standard weight-free evaluation of self-supervised features \cite{simeoni2025dinov3}: each test cell is labelled by the label fraction among its $k=20$ nearest training cells under cosine distance, on the same folds and buffer as the probe. AUC $0.90$ [0.87, 0.92] for DINOv3, $0.89$ [0.79, 0.91] for BenthicDINO and $0.77$ [0.72, 0.81] for ACF, against the probe's $0.93$, $0.92$ and $0.78$; its look sensitivity over 200 draws is $\pm 0.006$. The coral label is present in the distances as well as along the probe direction.
\paragraph{Look sensitivity of the coral AUC (Section~4.1).} With the fold models fixed (DINOv3, label threshold 0.5, 250\,m buffer): 200 random look draws $0.926$--$0.930$, choosing for every cell the look with the highest probability $0.93$, the lowest $0.93$, the mean of a cell's looks $0.93$, the look that most disfavours the probe (positives' lowest, negatives' highest) $0.89$, the look that most favours it $0.96$. 64\,\% of cells have one look and cannot change between draws.
\paragraph{Probe-direction splits and floors (Section~4.4).} The held-out ICC of the probe direction uses a two-fold 500\,m checkerboard split with a 100\,m buffer, the residual autocorrelation range, swapped and averaged (both training halves keep more than 900 coral cells). Without a buffer it is $0.65$ on 100 and 250\,m checkerboards and $0.61$ at 500\,m, so leakage inflates it slightly; with a 250\,m buffer it falls to $0.50$--$0.52$, but there one training half is left with as few as 64 coral cells, so part of that drop is a starved fit. The floors of Section~4.4 are probes fitted to labels permuted across cells with each cell's looks kept together (100 permutations), 1{,}000 random directions and the ten leading principal components, all evaluated on the same held-out cells.

\section{Uncertainty (Section~3.4)}
\paragraph{Reporting.} Every statistic is reported as a look-sensitivity over $R=200$ random draws of one look per cell (median and 2.5--97.5\,\% range) and over 1{,}000 replicates of a 250\,m spatial block bootstrap of the cells (63 blocks, 46 among the repeat-look cells). The intervals in the tables combine the two. The block exceeds the autocorrelation range of the probe residuals and is slightly below that of the embedding over the repeat-look cells (below).
\paragraph{Spatial autocorrelation ranges.} Practical ranges (95\,\% of the sill) of spherical / exponential variogram fits with nugget, median over five look draws, on an 8{,}000-cell subsample (scikit-gstat \cite{malicke2022scikitgstat}); Moran's $I$ \cite{moran1950notes} (esda, PySAL \cite{rey2007pysal}) on 25\,m distance rings. DINOv3 embedding (cosine), all cells: 743 / 689\,m, empirical plateau 488\,m, nugget share 0.69--0.79; repeat-look cells only: 283--299 / 258--274\,m. BenthicDINO: 284 / 244\,m (all), 161--182 / 146--156\,m (repeat-look). Probe residuals (label threshold 0.5): 103 / 89\,m, Moran's $I$ non-significant beyond 238\,m. Per-cell reliability $\rho_i$: 604 / 528\,m (DINOv3), 600 / 463\,m (BenthicDINO). Coral fraction: 289 / 290\,m. Depth: 1{,}232\,m spherical, no sill under the exponential model (a trend). Block counts at 100 / 250 / 500\,m: 318 / 63 / 23 over all cells and 223 / 46 / 16 over the repeat-look cells. The 250\,m probe buffer is beyond the residual range; the 250\,m bootstrap block exceeds the residual range but is slightly below the embedding range on the repeat-look cells and below the $\rho_i$ range, so intervals on the reliability rows may be somewhat optimistic; intervals at 100 and 500\,m blocks were $0.6$--$1.3\times$ those at 250\,m and no conclusion changed.

\section{Nearest-neighbour galleries (Sections~4.2 and 5)}
Each row is one cell: the query look (red), the same cell's other look (blue) with its rank among all looks of other cells, and the nearest neighbours excluding the cell and anything within 50\,m, with cosine distance, the cell's coral fraction and the distance from the query.
\begin{figure*}[p]\centering
\IfFileExists{figures/survey/supp/nn_galleries_dinov3_pm.jpg}{\includegraphics[width=\textwidth]{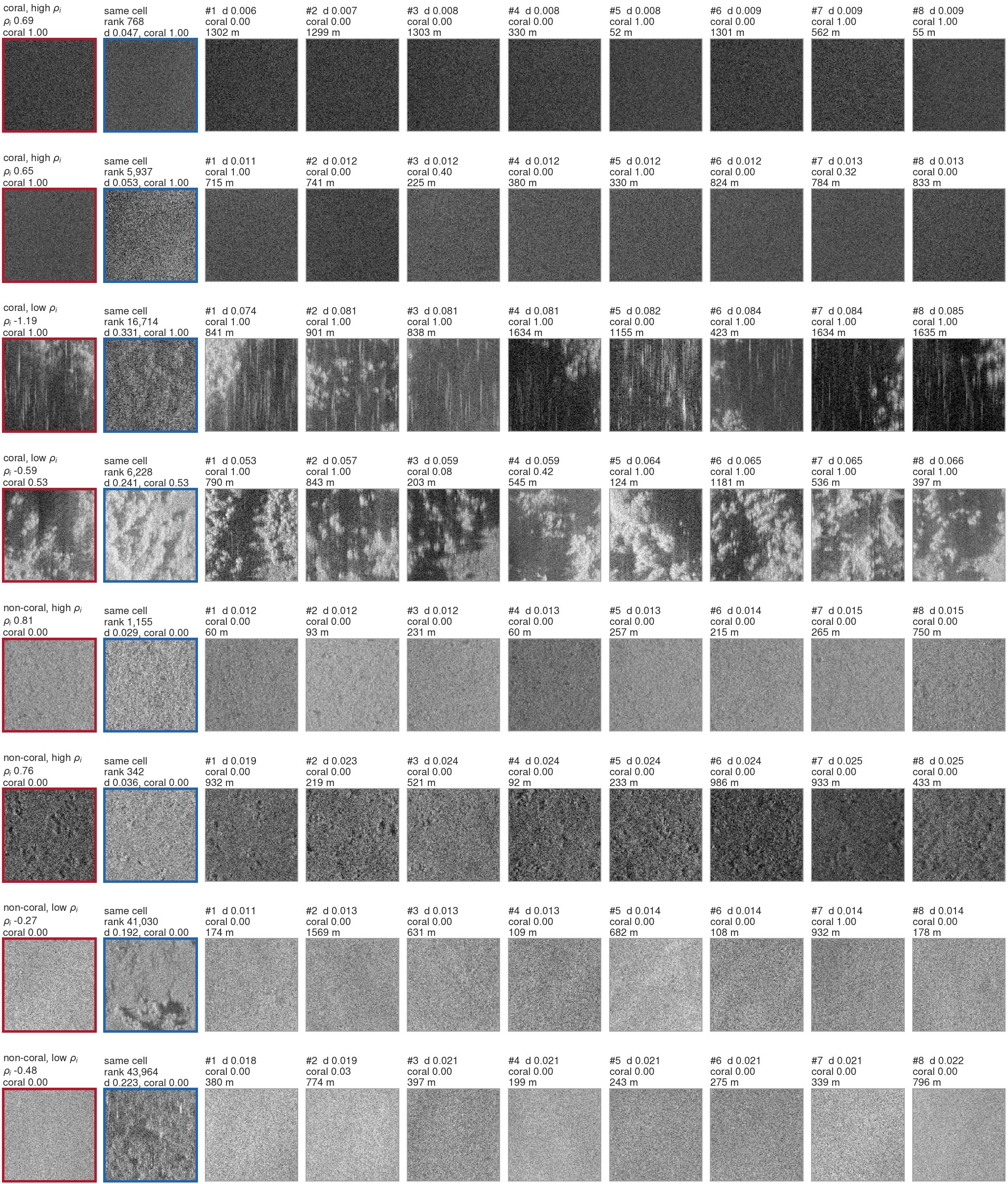}}{}
\caption{Nearest-neighbour galleries, DINOv3.}\label{fig:s-gal-dino}\end{figure*}
\begin{figure*}[p]\centering
\IfFileExists{figures/survey/supp/nn_galleries_benthicdino_faithful.jpg}{\includegraphics[width=\textwidth]{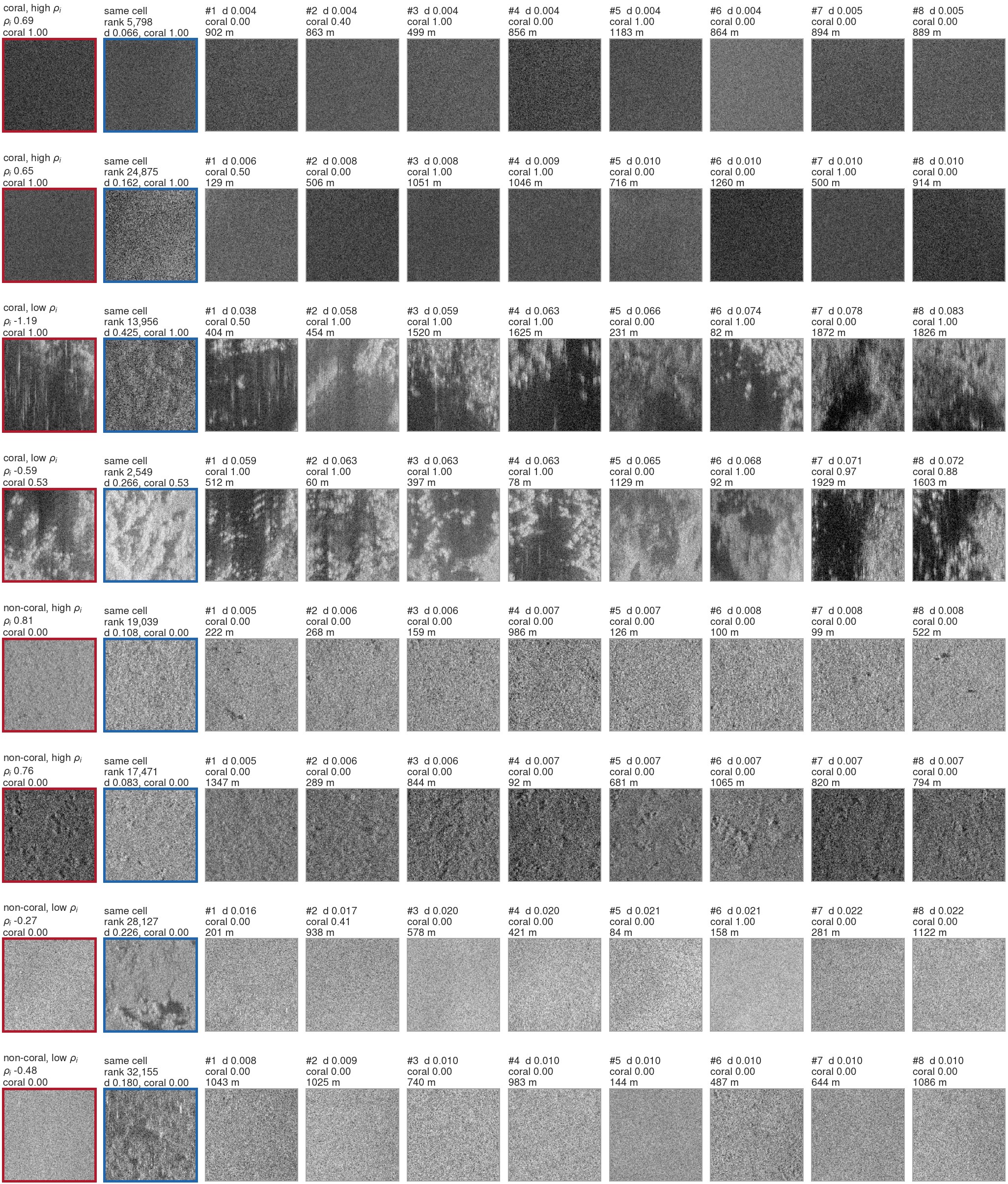}}{}
\caption{Nearest-neighbour galleries, BenthicDINO.}\label{fig:s-gal-bd}\end{figure*}
\begin{figure*}[p]\centering
\IfFileExists{figures/survey/supp/nn_galleries_acf.jpg}{\includegraphics[width=\textwidth]{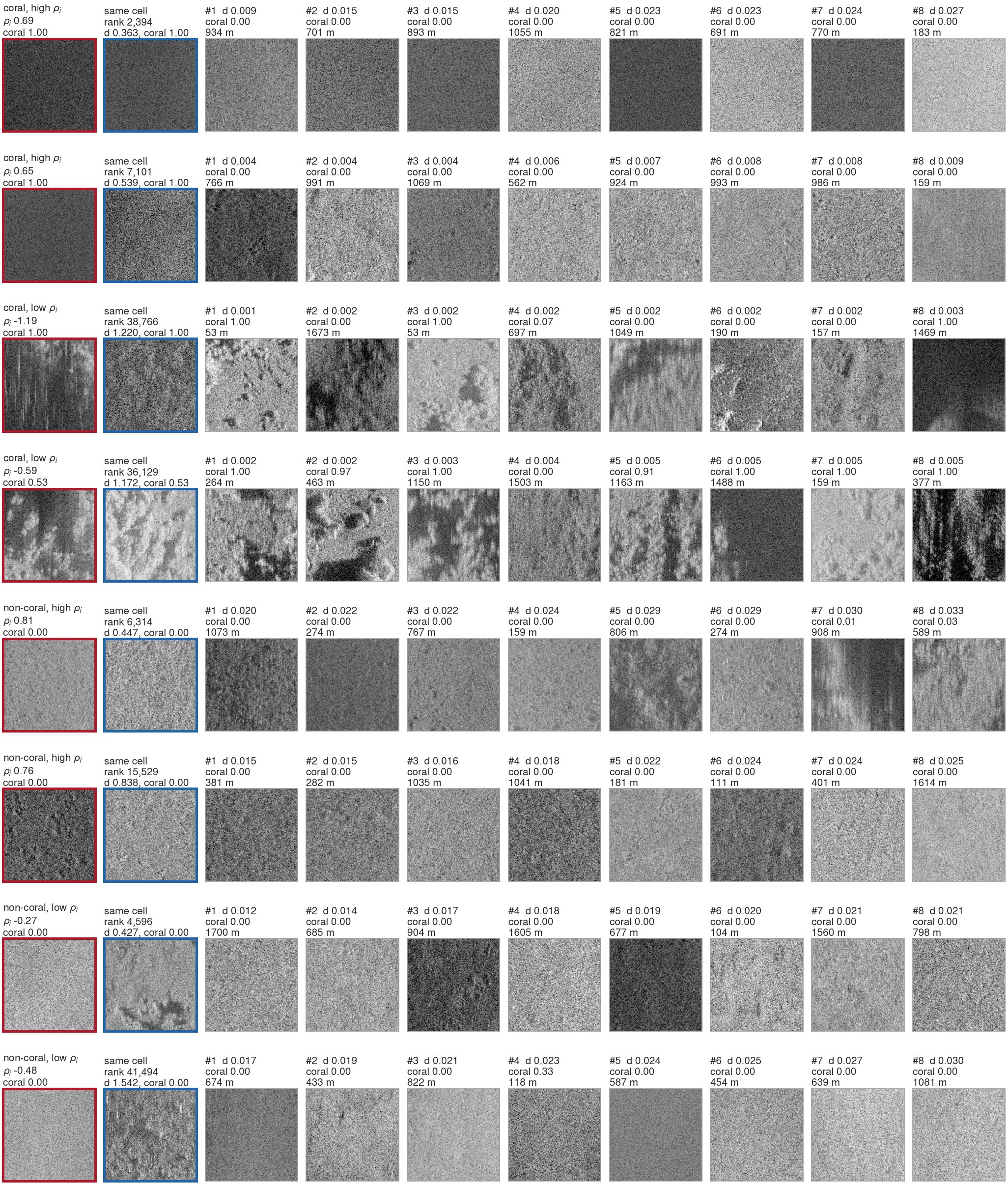}}{}
\caption{Nearest-neighbour galleries, ACF.}\label{fig:s-gal-acf}\end{figure*}